\documentclass{INTERSPEECH2023}
\pdfoutput=1

\interspeechcameraready

\usepackage{latexsym}
\usepackage{amsmath}
\usepackage{url}
\usepackage{amssymb}
\usepackage{amsfonts}
\usepackage{graphicx}
\usepackage{tabularx}
\usepackage{multirow}
\usepackage{arydshln}
\usepackage{mathtools,nccmath}
\usepackage[utf8]{inputenc}
\usepackage[T5]{fontenc}
\usepackage{enumitem}
\usepackage{todonotes}
\usepackage[most]{tcolorbox}
\usepackage[draft]{minted}
\usepackage[FIGTOPCAP]{subfigure}
\usepackage{helvet}  
\usepackage{courier}  
\usepackage{graphicx} 
\usepackage{algorithm}
\usepackage{algorithmic}

\makeatletter
\def\PYG@reset{\let\PYG@it=\relax \let\PYG@bf=\relax%
    \let\PYG@ul=\relax \let\PYG@tc=\relax%
    \let\PYG@bc=\relax \let\PYG@ff=\relax}
\def\PYG@tok#1{\csname PYG@tok@#1\endcsname}
\def\PYG@toks#1+{\ifx\relax#1\empty\else%
    \PYG@tok{#1}\expandafter\PYG@toks\fi}
\def\PYG@do#1{\PYG@bc{\PYG@tc{\PYG@ul{%
    \PYG@it{\PYG@bf{\PYG@ff{#1}}}}}}}
\def\PYG#1#2{\PYG@reset\PYG@toks#1+\relax+\PYG@do{#2}}

\expandafter\def\csname PYG@tok@w\endcsname{\def\PYG@tc##1{\textcolor[rgb]{0.73,0.73,0.73}{##1}}}
\expandafter\def\csname PYG@tok@c\endcsname{\let\PYG@it=\textit\def\PYG@tc##1{\textcolor[rgb]{0.25,0.50,0.50}{##1}}}
\expandafter\def\csname PYG@tok@cp\endcsname{\def\PYG@tc##1{\textcolor[rgb]{0.74,0.48,0.00}{##1}}}
\expandafter\def\csname PYG@tok@k\endcsname{\let\PYG@bf=\textbf\def\PYG@tc##1{\textcolor[rgb]{0.00,0.50,0.00}{##1}}}
\expandafter\def\csname PYG@tok@kp\endcsname{\def\PYG@tc##1{\textcolor[rgb]{0.00,0.50,0.00}{##1}}}
\expandafter\def\csname PYG@tok@kt\endcsname{\def\PYG@tc##1{\textcolor[rgb]{0.69,0.00,0.25}{##1}}}
\expandafter\def\csname PYG@tok@o\endcsname{\def\PYG@tc##1{\textcolor[rgb]{0.40,0.40,0.40}{##1}}}
\expandafter\def\csname PYG@tok@ow\endcsname{\let\PYG@bf=\textbf\def\PYG@tc##1{\textcolor[rgb]{0.67,0.13,1.00}{##1}}}
\expandafter\def\csname PYG@tok@nb\endcsname{\def\PYG@tc##1{\textcolor[rgb]{0.00,0.50,0.00}{##1}}}
\expandafter\def\csname PYG@tok@nf\endcsname{\def\PYG@tc##1{\textcolor[rgb]{0.00,0.00,1.00}{##1}}}
\expandafter\def\csname PYG@tok@nc\endcsname{\let\PYG@bf=\textbf\def\PYG@tc##1{\textcolor[rgb]{0.00,0.00,1.00}{##1}}}
\expandafter\def\csname PYG@tok@nn\endcsname{\let\PYG@bf=\textbf\def\PYG@tc##1{\textcolor[rgb]{0.00,0.00,1.00}{##1}}}
\expandafter\def\csname PYG@tok@ne\endcsname{\let\PYG@bf=\textbf\def\PYG@tc##1{\textcolor[rgb]{0.82,0.25,0.23}{##1}}}
\expandafter\def\csname PYG@tok@nv\endcsname{\def\PYG@tc##1{\textcolor[rgb]{0.10,0.09,0.49}{##1}}}
\expandafter\def\csname PYG@tok@no\endcsname{\def\PYG@tc##1{\textcolor[rgb]{0.53,0.00,0.00}{##1}}}
\expandafter\def\csname PYG@tok@nl\endcsname{\def\PYG@tc##1{\textcolor[rgb]{0.63,0.63,0.00}{##1}}}
\expandafter\def\csname PYG@tok@ni\endcsname{\let\PYG@bf=\textbf\def\PYG@tc##1{\textcolor[rgb]{0.60,0.60,0.60}{##1}}}
\expandafter\def\csname PYG@tok@na\endcsname{\def\PYG@tc##1{\textcolor[rgb]{0.49,0.56,0.16}{##1}}}
\expandafter\def\csname PYG@tok@nt\endcsname{\let\PYG@bf=\textbf\def\PYG@tc##1{\textcolor[rgb]{0.00,0.50,0.00}{##1}}}
\expandafter\def\csname PYG@tok@nd\endcsname{\def\PYG@tc##1{\textcolor[rgb]{0.67,0.13,1.00}{##1}}}
\expandafter\def\csname PYG@tok@s\endcsname{\def\PYG@tc##1{\textcolor[rgb]{0.73,0.13,0.13}{##1}}}
\expandafter\def\csname PYG@tok@sd\endcsname{\let\PYG@it=\textit\def\PYG@tc##1{\textcolor[rgb]{0.73,0.13,0.13}{##1}}}
\expandafter\def\csname PYG@tok@si\endcsname{\let\PYG@bf=\textbf\def\PYG@tc##1{\textcolor[rgb]{0.73,0.40,0.53}{##1}}}
\expandafter\def\csname PYG@tok@se\endcsname{\let\PYG@bf=\textbf\def\PYG@tc##1{\textcolor[rgb]{0.73,0.40,0.13}{##1}}}
\expandafter\def\csname PYG@tok@sr\endcsname{\def\PYG@tc##1{\textcolor[rgb]{0.73,0.40,0.53}{##1}}}
\expandafter\def\csname PYG@tok@ss\endcsname{\def\PYG@tc##1{\textcolor[rgb]{0.10,0.09,0.49}{##1}}}
\expandafter\def\csname PYG@tok@sx\endcsname{\def\PYG@tc##1{\textcolor[rgb]{0.00,0.50,0.00}{##1}}}
\expandafter\def\csname PYG@tok@m\endcsname{\def\PYG@tc##1{\textcolor[rgb]{0.40,0.40,0.40}{##1}}}
\expandafter\def\csname PYG@tok@gh\endcsname{\let\PYG@bf=\textbf\def\PYG@tc##1{\textcolor[rgb]{0.00,0.00,0.50}{##1}}}
\expandafter\def\csname PYG@tok@gu\endcsname{\let\PYG@bf=\textbf\def\PYG@tc##1{\textcolor[rgb]{0.50,0.00,0.50}{##1}}}
\expandafter\def\csname PYG@tok@gd\endcsname{\def\PYG@tc##1{\textcolor[rgb]{0.63,0.00,0.00}{##1}}}
\expandafter\def\csname PYG@tok@gi\endcsname{\def\PYG@tc##1{\textcolor[rgb]{0.00,0.63,0.00}{##1}}}
\expandafter\def\csname PYG@tok@gr\endcsname{\def\PYG@tc##1{\textcolor[rgb]{1.00,0.00,0.00}{##1}}}
\expandafter\def\csname PYG@tok@ge\endcsname{\let\PYG@it=\textit}
\expandafter\def\csname PYG@tok@gs\endcsname{\let\PYG@bf=\textbf}
\expandafter\def\csname PYG@tok@gp\endcsname{\let\PYG@bf=\textbf\def\PYG@tc##1{\textcolor[rgb]{0.00,0.00,0.50}{##1}}}
\expandafter\def\csname PYG@tok@ga\endcsname{\def\PYG@tc##1{\textcolor[rgb]{0.50,0.00,0.50}{##1}}}
\expandafter\def\csname PYG@tok@go\endcsname{\def\PYG@tc##1{\textcolor[rgb]{0.53,0.53,0.53}{##1}}}
\expandafter\def\csname PYG@tok@gt\endcsname{\def\PYG@tc##1{\textcolor[rgb]{0.00,0.27,0.87}{##1}}}
\expandafter\def\csname PYG@tok@err\endcsname{\def\PYG@bc##1{\setlength{\fboxsep}{0pt}\fcolorbox[rgb]{1.00,0.00,0.00}{1,1,1}{\strut ##1}}}
\expandafter\def\csname PYG@tok@kc\endcsname{\let\PYG@bf=\textbf\def\PYG@tc##1{\textcolor[rgb]{0.00,0.50,0.00}{##1}}}
\expandafter\def\csname PYG@tok@kd\endcsname{\let\PYG@bf=\textbf\def\PYG@tc##1{\textcolor[rgb]{0.00,0.50,0.00}{##1}}}
\expandafter\def\csname PYG@tok@kn\endcsname{\let\PYG@bf=\textbf\def\PYG@tc##1{\textcolor[rgb]{0.00,0.50,0.00}{##1}}}
\expandafter\def\csname PYG@tok@kr\endcsname{\let\PYG@bf=\textbf\def\PYG@tc##1{\textcolor[rgb]{0.00,0.50,0.00}{##1}}}
\expandafter\def\csname PYG@tok@bp\endcsname{\def\PYG@tc##1{\textcolor[rgb]{0.00,0.50,0.00}{##1}}}
\expandafter\def\csname PYG@tok@fm\endcsname{\def\PYG@tc##1{\textcolor[rgb]{0.00,0.00,1.00}{##1}}}
\expandafter\def\csname PYG@tok@vc\endcsname{\def\PYG@tc##1{\textcolor[rgb]{0.10,0.09,0.49}{##1}}}
\expandafter\def\csname PYG@tok@vg\endcsname{\def\PYG@tc##1{\textcolor[rgb]{0.10,0.09,0.49}{##1}}}
\expandafter\def\csname PYG@tok@vi\endcsname{\def\PYG@tc##1{\textcolor[rgb]{0.10,0.09,0.49}{##1}}}
\expandafter\def\csname PYG@tok@vm\endcsname{\def\PYG@tc##1{\textcolor[rgb]{0.10,0.09,0.49}{##1}}}
\expandafter\def\csname PYG@tok@sa\endcsname{\def\PYG@tc##1{\textcolor[rgb]{0.73,0.13,0.13}{##1}}}
\expandafter\def\csname PYG@tok@sb\endcsname{\def\PYG@tc##1{\textcolor[rgb]{0.73,0.13,0.13}{##1}}}
\expandafter\def\csname PYG@tok@sc\endcsname{\def\PYG@tc##1{\textcolor[rgb]{0.73,0.13,0.13}{##1}}}
\expandafter\def\csname PYG@tok@dl\endcsname{\def\PYG@tc##1{\textcolor[rgb]{0.73,0.13,0.13}{##1}}}
\expandafter\def\csname PYG@tok@s2\endcsname{\def\PYG@tc##1{\textcolor[rgb]{0.73,0.13,0.13}{##1}}}
\expandafter\def\csname PYG@tok@sh\endcsname{\def\PYG@tc##1{\textcolor[rgb]{0.73,0.13,0.13}{##1}}}
\expandafter\def\csname PYG@tok@s1\endcsname{\def\PYG@tc##1{\textcolor[rgb]{0.73,0.13,0.13}{##1}}}
\expandafter\def\csname PYG@tok@mb\endcsname{\def\PYG@tc##1{\textcolor[rgb]{0.40,0.40,0.40}{##1}}}
\expandafter\def\csname PYG@tok@mf\endcsname{\def\PYG@tc##1{\textcolor[rgb]{0.40,0.40,0.40}{##1}}}
\expandafter\def\csname PYG@tok@mh\endcsname{\def\PYG@tc##1{\textcolor[rgb]{0.40,0.40,0.40}{##1}}}
\expandafter\def\csname PYG@tok@mi\endcsname{\def\PYG@tc##1{\textcolor[rgb]{0.40,0.40,0.40}{##1}}}
\expandafter\def\csname PYG@tok@il\endcsname{\def\PYG@tc##1{\textcolor[rgb]{0.40,0.40,0.40}{##1}}}
\expandafter\def\csname PYG@tok@mo\endcsname{\def\PYG@tc##1{\textcolor[rgb]{0.40,0.40,0.40}{##1}}}
\expandafter\def\csname PYG@tok@ch\endcsname{\let\PYG@it=\textit\def\PYG@tc##1{\textcolor[rgb]{0.25,0.50,0.50}{##1}}}
\expandafter\def\csname PYG@tok@cm\endcsname{\let\PYG@it=\textit\def\PYG@tc##1{\textcolor[rgb]{0.25,0.50,0.50}{##1}}}
\expandafter\def\csname PYG@tok@cpf\endcsname{\let\PYG@it=\textit\def\PYG@tc##1{\textcolor[rgb]{0.25,0.50,0.50}{##1}}}
\expandafter\def\csname PYG@tok@c1\endcsname{\let\PYG@it=\textit\def\PYG@tc##1{\textcolor[rgb]{0.25,0.50,0.50}{##1}}}
\expandafter\def\csname PYG@tok@cs\endcsname{\let\PYG@it=\textit\def\PYG@tc##1{\textcolor[rgb]{0.25,0.50,0.50}{##1}}}

\def\PYGZus{\char`\_}

\def\PYGZas{$\ast$}

\makeatother

\title{XPhoneBERT: A Pre-trained Multilingual Model for Phoneme Representations for Text-to-Speech}

\name{Linh The Nguyen, Thinh Pham, Dat Quoc Nguyen}
\address{VinAI Research, Vietnam}
\email{\{v.linhnt140, v.thinhphp1, v.datnq9\}@vinai.io}

\begin{document}
\maketitle

\begin{abstract}
We present XPhoneBERT, the first multilingual model pre-trained to learn phoneme representations for the downstream text-to-speech (TTS) task. Our XPhoneBERT has the same model architecture as BERT-base, trained using the RoBERTa pre-training approach on 330M phoneme-level sentences from nearly 100 languages and locales. Experimental results show that employing XPhoneBERT as an input phoneme encoder significantly boosts the performance of a strong neural TTS model in terms of naturalness and prosody and also helps produce fairly high-quality speech with limited training data. We publicly release our pre-trained XPhoneBERT with the hope that it would facilitate future research and downstream TTS applications for multiple languages.
\end{abstract}

\medskip


\section{Introduction}
\label{sec:intro}

Advancements in neural TTS technology have led to significant improvements in producing natural-sounding speech \cite{kong2020hifi, ren2019fastspeech, popov2021grad, tan2022naturalspeech}, increasingly closing the gap between artificial speech and human-recorded speech in terms of naturalness. 
Early work such as \cite{shen2018natural} employs an encoder to directly convert input raw texts to mel-spectrograms that are then fed into a decoder to generate output speech. Other works often take phoneme sequences as input for their encoder \cite{li2019neural,kim2020glow, renfastspeech, kim2021conditional}. Here, the encoder in these works might be extended by utilizing recent large-scale pre-trained language models that are learned from unlabeled textual or phonemic description data to enhance the naturalness of speech outputs.

The large-scale pre-trained language models, e.g. BERT \cite{devlin-etal-2019-bert}, RoBERTa \cite{RoBERTa} and ALBERT \cite{LanCGGSS20}, have proved their effectiveness, improving state-of-the-art performances of various natural language processing research and application tasks. 
For TTS, some works incorporate contextualized word embeddings generated by the pre-trained BERT \cite{devlin-etal-2019-bert} into their standard encoder \cite{hayashi19_interspeech, kenter2020improving,xu2021improving}. In general, an input phoneme sequence is fed into the standard TTS encoder to produce phoneme representations, while its corresponding raw text is fed into BERT to obtain contextualized word embeddings. To construct the
input vectors of the TTS decoder, the produced representations of the input phonemes are concatenated with the BERT-based contextualized embedding of the corresponding word that the phonemes belong to. As a result, BERT helps increase the quality of the output synthesized speech. Here, the pre-trained BERT is used to provide additional contextual information for phoneme representations indirectly. Therefore, it might be better if the contextualized phoneme representations are directly produced by a pre-trained BERT-type model that is learned from unlabeled phoneme-level data. 

Recent works confirm that pre-trained models for phoneme representations, including PnG BERT \cite{jia2021png}, Mixed-Phoneme BERT \cite{zhang2022mixed} and Phoneme-level BERT \cite{phoneme-levelBERT},  help improve advanced TTS systems. PnG BERT and Mixed-Phoneme BERT are trained based on the BERT pre-training approach \cite{devlin-etal-2019-bert}, in which PnG BERT takes both phonemes and graphemes (i.e. subword tokens) as the input, while Mixed-Phoneme BERT  takes both phonemes and sup-phoneme tokens as the input. Phoneme-level BERT is trained based on the ALBERT pre-training approach \cite{LanCGGSS20}, only taking phonemes as the input. In addition to the standard masked token prediction task as used in PnG BERT and Mixed-Phoneme BERT, the Phoneme-level BERT also proposes an additional auxiliary task that predicts the corresponding grapheme for each phoneme. Here, PnG BERT, Mixed-Phoneme BERT and Phoneme-level BERT can be directly used as an input encoder in a typical neural TTS system. Note that the success of these pre-trained language models has been limited to the English language only. Taking into account a societal, linguistic, cultural, machine learning and cognitive perspective \cite{donlpotherlanguages}, it is worth exploring pre-trained models for phoneme representations in languages other than English. 

To fill the gap, we train the \textit{first} large-scale multilingual language model for phoneme representations, using a pre-training corpus of 330M phonemic description sentences from nearly 100 languages and locales.  Our model is trained based on the RoBERTa pre-training approach \cite{RoBERTa}, using the BERT-base model configuration \cite{devlin-etal-2019-bert}. We conduct experiments on the downstream TTS task, directly employing our model as an input phoneme encoder of the strong model VITS \cite{kim2021conditional}. Experimental results show that our model helps boost the performance of VITS, obtaining more natural prosody than the original VITS without pre-training and also producing fairly high-quality synthesized speech with limited training data. We summarize our contribution as follows:

\begin{itemize}

\item We present the first large-scale pre-trained multilingual model for phoneme representations, which we name XPhoneBERT.

\item On the downstream TTS task, XPhoneBERT helps significantly improve the performance of the strong baseline VITS, thus confirming its  effectiveness.

\item We publicly release XPhoneBERT at \url{https://github.com/VinAIResearch/XPhoneBERT}. We hope that our XPhoneBERT model would help facilitate future research and downstream TTS applications for nearly 100 languages and locales.

\end{itemize}

\section{Our XPhoneBERT}
\label{sec:mphone}

This section outlines the architecture and describes the multilingual pre-training corpus and optimization setup that we use for XPhoneBERT. 

\subsection{Model architecture}

XPhoneBERT has the same model architecture as BERT-base \cite{devlin-etal-2019-bert}---a multi-layer bidirectional Transformer encoder \cite{NIPS2017_7181}---in which the number of Transformer blocks, the hidden size and the number of self-attention heads are 12, 768 and 12, respectively. To pre-train XPhoneBERT, we use the masked language modeling objective \cite{devlin-etal-2019-bert} and follow the RoBERTa pre-training approach \cite{RoBERTa}  which robustly optimizes BERT  for better performance, i.e. using a dynamic masking strategy and without the next sentence prediction objective. Given the popularity of BERT and RoBERTa, we do not further detail about the architecture here.  See \cite{devlin-etal-2019-bert,RoBERTa} for more information.

\subsection{Multilingual pre-training data}

Our multilingual pre-training dataset is constructed following three phases. The first phase is to collect text documents and then perform word and sentence segmentation as well as duplicate removal and text normalization. The second phase is to convert texts into phonemes, employing the CharsiuG2P toolkit \cite{zhu2022charsiu-g2p} that supports 90+ languages and locales. Finally, the third phase is to perform phoneme segmentation.

\subsubsection{First phase: Data collection and pre-processing}

We collect texts for the languages supported by CharsiuG2P. Here, we employ the multilingual datasets \texttt{wiki40b} \cite{guo-etal-2020-wiki} and \texttt{wikipedia}  \cite{wikidump}, available to download from the Hugging Face \textit{datasets} library \cite{lhoest-etal-2021-datasets}. 
In particular, we first download the \texttt{wiki40b} dataset consisting of text documents for 41 Wikipedia languages and locales.\footnote{\url{https://huggingface.co/datasets/wiki40b}} We then  use \texttt{wikipedia} to extract texts from Wikipedia dumps for remaining languages other than those belonging to \texttt{wiki40b}.\footnote{\url{https://huggingface.co/datasets/wikipedia}} 

We perform word and sentence segmentation on all text documents in each language by  using the \texttt{spaCy} toolkit,\footnote{\url{https://spacy.io}} except for Vietnamese where we employ RDRSegmenter \cite{nguyen-etal-2018-fast} from the VnCoreNLP toolkit \cite{vu-etal-2018-vncorenlp}. We then lowercase all sentences and filter out duplicate sentences and single-word ones.  We also apply text normalization to convert texts from their written form into their verbalized form for only English, German, Spanish, Vietnamese and Chinese (it is because we could not find an effective text normalization tool publicly available for other languages). 
Here, we use the text normalization component from the NVIDIA NeMo toolkit \cite{nemo} for English, German, Spanish and Chinese, and the \texttt{Vinorm} text normalization package for Vietnamese.\footnote{\url{https://github.com/v-nhandt21/Vinorm}}  

\begin{table}[!t]
\centering
\caption{Our pre-training data statistics. ``LCode'' denotes the ISO 639-3 code for each language or locale, while ``\#s'' denotes the number of sentences.}
\resizebox{7.5cm}{!}{
\begin{tabular}{l|l|l|l|l|l}
\hline
\textbf{LCode} & \textbf{\#s (K)} & \textbf{LCode} & \textbf{\#s (K)} & \textbf{LCode} & \textbf{\#s (K)} \\ 
\hline
ady & 2 & glg & 3793 & ron & 1816 \\ \hline
afr & 1793 & grc & 947 & rus & 15923 \\ \hline
amh & 73 & gre & 947 & san & 114 \\ \hline
ara & 2820 & grn & 60 & slo & 1143 \\ \hline
arg & 383 & guj & 211 & slv & 1167 \\ \hline
arm-e & 2989 & hbs-cyrl & 2007 & sme & 27 \\ \hline
arm-w & 175 & hbs-latn & 2007 & snd & 215 \\ \hline
aze & 3139 & hin & 287 & spa & 3936 \\ \hline
bak & 1272 & hun & 4372 & spa-latin & 3936 \\ \hline
bel & 2750 & ice & 776 & spa-me & 3936 \\ \hline
ben & 1785 & ido & 224 & sqi & 1373 \\ \hline
bos & 1464 & ina & 100 & srp & 2449 \\ \hline
bul & 1919 & ind & 2196 & swa & 537 \\ \hline
bur & 393 & ita & 12335 & swe & 5226 \\ \hline
cat & 4017 & jam & 8 & tam & 2289 \\ \hline
cze & 4542 & jpn & 12197 & tat & 984 \\ \hline
dan & 1714 & kaz & 1850 & tgl & 628 \\ \hline
dut & 7683 & khm & 93 & tha & 567 \\ \hline
egy & 3093 & kor & 2384 & tts & 567 \\ \hline
eng-uk & 33515 & kur & 335 & tuk & 105 \\ \hline
eng-us & 33515 & lat-clas & 597 & tur & 2148 \\ \hline
epo & 4333 & lat-eccl & 597 & ukr & 6967 \\ \hline
est & 1558 & lit & 1087 & vie-c & 2519 \\ \hline
eus & 3429 & ltz & 817 & vie-n & 2519 \\ \hline
fas & 1957 & mac & 2597 & vie-s & 2519 \\ \hline
fin & 4100 & min & 377 & wel-nw & 714 \\ \hline
fra & 11255 & mlt & 180 & wel-sw & 714 \\ \hline
fra-qu & 11255 & ori & 158 & yue & 908 \\ \hline
geo & 1211 & pap & 27 & zho-s & 6934 \\ \hline
ger & 33845 & pol & 7045 & zho-t & 6955 \\ \hline
gla & 121 & por-bz & 3437 & \_ & \_ \\ \hline 
gle & 488 & por-po & 3437 & \_ & \_  \\ \hline
\end{tabular}
}
\label{tab:data_stat}
\end{table}

\begin{figure*}[!ht]
\begin{Verbatim}[commandchars=\\\{\}]
\PYG{k+kn}{from} {transformers} \PYG{k+kn}{import} \PYG{l+s+s2}{AutoModel}, \PYG{l+s+s2}{AutoTokenizer}

tokenizer = \PYG{l+s+s2}{AutoTokenizer}.\PYG{ga}{from\PYGZus{}pretrained}(\PYG{vi}{"vinai/xphonebert-base"})
model = \PYG{l+s+s2}{AutoModel}.\PYG{ga}{from\PYGZus{}pretrained}(\PYG{vi}{"vinai/xphonebert-base"})
input\PYGZus{}phonemes = \PYG{vi}{"{\textprimstress}e \textipa{I} \textbf{\_} \s{}m \textschwa \textltilde t i {\textprimstress}\textltilde \textipa{I} \textipa{\ng} w \textschwa \textltilde \textbf{\_} {\textprimstress}m  \textscripta d \textschwa \textltilde"} 
input\PYGZus{}ids = \PYG{ga}{tokenizer}(input\PYGZus{}phonemes, return\PYGZus{}tensors=\PYG{vi}{"pt"})
features = \PYG{ga}{model}(\PYGZas{}\PYGZas{}input\PYGZus{}ids)
\end{Verbatim}
\caption{An example code using XPhoneBERT for feature extraction with the Hugging Face \texttt{transformers} library in Python. Here, the input phonemes represent  a phonemic description of the word-level sequence ``a multilingual model''.}
\label{fig:code}
\end{figure*}

\subsubsection{Second phase: Text-to-phoneme conversion}\label{sssec:phase2}

For each language whose locales do not have their own Wikipedia data,\footnote{Languages whose locales do not have their own Wikipedia data are: English (eng-uk \& eng-us), French (fra \& fra-qu), Greek (grc \& gre), Latin (lat-clas \& lat-eccl), Portuguese (por-po \& por-bz), Serbo-Croatian (hbs-latn \& hbs-cyrl), Spanish (spa \& spa-latin \& spa-me), Thai (tha \& tts), Vietnamese (vie-n, vie-c \& vie-s) and Welsh (wel-nw \& wel-sw). By contrast, Armenian and Chinese have the corresponding Wikipedia data for their locales (Armenian: arm-e \& arm-w; Chinese: min, yue, zho-s \& zho-t).} we randomly divide the language's Wikipedia data into equal parts (each with the same number of sentences), with each part corresponding to a locale. For example, we divide 67 million English sentences into two equal parts that are then separately converted into phonemic descriptions in British English (\texttt{eng-uk}) and American English (\texttt{eng-us}). 

To convert sentences into their phonemic description, we employ the grapheme-to-phoneme conversion toolkit  CharsiuG2P \cite{zhu2022charsiu-g2p}. The pre-trained CharsiuG2P is a strong multilingual Transformer-based model that generates the pronunciation of a word given its orthographic form and ISO 639-3 language code pair. Following the recommendation from \cite{zhu2022charsiu-g2p}, if the input word is in the CharsiuG2P toolkit's pronunciation dictionary of the target language/locale, we employ the pronunciation dictionary to generate the word's phonemic description. Otherwise, if the word is out of the vocabulary, we employ the pre-trained CharsiuG2P model to generate its phonemic description. 

For example, given an input word ``model'' and the language code  \texttt{eng-us} of American English, CharsiuG2P produces an output phoneme sequence of ``{\textprimstress}m{\textscripta}{d}{\textschwa}{\textltilde}''. Such an American English sentence as ``a multilingual model'' is thus converted into  a phoneme sequence of ``{\textprimstress}e\textipa{I}\ \s{}m\textschwa\textltilde{ti}{\textprimstress}\textltilde\textipa{I}\textipa{\ng}w\textschwa\textltilde\ {\textprimstress}m\textscripta{d}\textschwa\textltilde''. 
Note that in this conversion phase, we keep punctuations intact, as do the TTS systems \cite{li2019neural,kim2020glow, renfastspeech, kim2021conditional}.

\subsubsection{Third phase: Phoneme segmentation}

CharsiuG2P converts each input word into a sequence of consecutive phonemes without a phoneme boundary indicator (e.g. white space). To better map between phonemes and speech \cite{kim2020glow,kim2021conditional,phoneme-levelBERT}, we would have to perform phoneme segmentation on the CharsiuG2P's output. Following \cite{Bernard2021}, we employ the \texttt{segments} toolkit for  phoneme segmentation.\footnote{\url{https://pypi.org/project/segments}} Thus an input word is now converted into a sequence of phonemes separated by white spaces, e.g. ``model'' is converted into ``{\textprimstress}m {\textscripta} {d} {\textschwa} {\textltilde}'' in \texttt{eng-us}. Since we use the white space to separate phonemes, to distinguish phonemes belonging to different word tokens, we employ a meta symbol $\mathbf{\_}$ (U+2581) for marking word boundaries. For example, the American English sentence  ``a multilingual model'' is now converted into the phoneme-segmented sequence ``{\textprimstress}e\ \textipa{I}\ \textbf{\_}\ \ \s{}m\ \textschwa\ \textltilde\ t\ i\ {\textprimstress}\textltilde\ \textipa{I}\ \textipa{\ng}\ w\ \textschwa\ \textltilde\ \textbf{\_}\ {\textprimstress}m\ \textscripta\ d\ \textschwa\ \textltilde''.\footnote{For convenience, we also create a Python package named  \texttt{text2phonemesequence}, incorporating both CharsiuG2P and \texttt{segments}, to perform a direct conversion from an input word-level sentence (e.g.  ``a multilingual model'') to an output phoneme-segmented sequence (e.g. ``{\textprimstress}e\ \textipa{I}\ \textbf{\_}\ \ \s{}m\ \textschwa\ \textltilde\ t\ i\ {\textprimstress}\textltilde\ \textipa{I}\ \textipa{\ng}\ w\ \textschwa\ \textltilde\ \textbf{\_}\ {\textprimstress}m\ \textscripta\ d\ \textschwa\ \textltilde'').}

\subsubsection{Pre-training data statistics}

Through the 3-phase construction process, we finally obtain a pre-training corpus of 330M phoneme-level sentences across 94 languages and locales. We present the data statistic for each language or locale in Table \ref{tab:data_stat}.
 
\subsection{Optimization}

We employ a white-space tokenizer, resulting in a vocabulary of 1960 phoneme types. Our XPhoneBERT thus has a total of 87.6M parameters. For training XPhoneBERT on our multilingual pre-training corpus, we employ the RoBERTa implementation \cite{RoBERTa} from the \texttt{fairseq} library \cite{ott2019fairseq}. We set a maximum sequence length of 512. We optimize the model using Adam \cite{KingmaB14} and use a batch size of 1024 sequence blocks across 8 A100 GPUs (40GB each) and a peak learning rate of 0.0001. We train for 20 epochs in about 18 days (here, the first 2 epochs are used for warming up the learning rate).

\subsection{Usage example}

To show the potential use for downstream tasks, we present in Figure \ref{fig:code} a basic usage of our pre-trained  model XPhoneBERT for feature extraction with the \texttt{transformers} library \cite{wolf-etal-2020-transformers}. More usage examples of XPhoneBERT can be found at the XPhoneBERT's GitHub repository.


\section{Experimental setup}
\label{sec:setup}

We evaluate the effectiveness of XPhoneBERT on the downstream text-to-speech (TTS) task. Due to a limited resource of human raters, we perform this TTS task for American English (\texttt{eng-us}) and Northern Vietnamese (\texttt{vie-n}).\footnote{The model weights of PnG BERT (\url{https://google.github.io/tacotron/publications/png_bert}), Mixed-Phoneme BERT (\url{https://speechresearch.github.io/mpbert}) and Phoneme-level BERT (\url{https://github.com/yl4579/PL-BERT}) are not published at the time of our empirical investigation (here, these pre-trained models are still not yet publicly available on 8th March 2023---the INTERSPEECH 2023’s paper update deadline). 
Therefore, we could not compare our multilingual XPhoneBERT with those monolingual models for English.}

\subsection{TTS datasets}

For English, we use the benchmark dataset LJSpeech \cite{ljspeech17} consisting of 13,100 audio clips of a single speaker with a total duration of about 24 hours (here, each clip is also provided with a gold-standard text transcription). 
Following \cite{kim2021conditional}, the dataset is split into training, validation and test sets of 12,500, 100 and 500 clip samples, respectively. 

For Vietnamese, we randomly sample 12,300 different medium-length sentences from the PhoBERT pre-training news data \cite{phobert}. We hire a professional speaker to read each sentence in a studio and record the corresponding audio, resulting in a total duration of about 18 hours for 12,300 high-quality audio clips. We split our Vietnamese TTS dataset into training, validation and test sets of 12,000, 100 and 200 clips, respectively.

\subsection{TTS modeling and training}

We employ the strong TTS model VITS \cite{kim2021conditional}.\footnote{\url{https://github.com/jaywalnut310/vits}} 
VITS is an end-to-end model that contains a Transformer encoder \cite{NIPS2017_7181} to encode the input phoneme sequence. We extend VITS with  XPhoneBERT by replacing the VITS's Transformer encoder with XPhoneBERT.

For the first setting of using the whole TTS training set, we train the original VITS model with optimal hyper-parameters used in its paper \cite{kim2021conditional}, e.g. using the AdamW optimizer \cite{loshchilov2018decoupled} with $\beta_1=0.8, \beta_2=0.99$ and the weight decay $\lambda=0.01$, and an initial learning rate of $2\times10^{-4}$ (here, the learning rate decay is scheduled by a $0.999^{1/8}$ factor in every epoch). We run for 300K training steps with a batch size of 64 (i.e. equivalent to about 1600 training epochs for both English and Vietnamese). For training the VITS variant extended with XPhoneBERT, we apply the same training protocol used for the original VITS. Here, XPhoneBERT is frozen in the first 25\% of the training steps and then updated during the remaining training steps.  

We also experiment with another setting where the TTS training data is limited. In particular, for each language, we randomly sample 5\% of the training audio clips, and then only use those sampled audios for training (total duration of about 1.2 hours for English and about 0.9 hours for Vietnamese). We apply the same training protocol used for the first setting with an exception that we run for 100K training steps.

\subsection{Evaluation protocol}
We evaluate the performance of TTS models using subjective  and objective metrics.
For subjective evaluation of the naturalness, for each language, following \cite{ren2019fastspeech,kim2020glow,kim2021conditional}, we randomly select 50 ground truth test audios and their text transcription to measure the Mean Opinion Score (MOS). Here, for each text transcription, we synthesize speeches using 4 different models (including the baseline VITS and the VITS variant extended with our XPhoneBERT, which are trained under the two different experimental settings detailed in the previous subsection). For each language, we hire 10 native speakers to rate each of the five speeches (i.e. the four synthesized speeches and the ground truth speech) on a naturalness scale from 1 to 5 with 1-point increments. Here, each rater does not know which model produces which speech. 

For objective evaluations of the distortion and intonation difference between the ground truth speech and the synthesized speech, we compute 
two  metrics of the mel-cesptrum distance (MCD; dB) and the F0 root mean square error (RMSE$_{\text{F}_\text{0}}$; cent), according to the implementation from \cite{tamamori2017speaker}.

\begin{table}[!t]
\centering
\caption{Obtained results on the English test set. ``100\%'' and ``5\%'' denote the first experimental setting of using the whole TTS training set and the second experimental setting of using only 5\% of the TTS training set for training, respectively. ``XPB'' abbreviates our XPhoneBERT. The MOS is reported with 95\% confidence intervals (here, each MOS score difference between two models is significant with p-value $<$ 0.05).}
\setlength{\tabcolsep}{0.3em}
\begin{tabular}{r|r|c|c|c}
\hline
\multicolumn{2}{c|}{\textbf{Model}} & \textbf{MOS} ($\uparrow$) & \textbf{MCD} ($\downarrow$) & \textbf{RMSE$_{\text{F}_\text{0}}$} ($\downarrow$)\\
\hline
\multicolumn{2}{r|}{Ground truth} & 4.39 $\pm$ 0.08 & 0.00 & 0.00\\
\hline
\hline
\multirow{2}{*}{\rotatebox[origin=c]{90}{100\%}}
& Baseline VITS & 4.00 $\pm$ 0.08 & 7.04 & 377 \\
\cline{2-5}
& VITS w/ XPB & \textbf{4.14} $\pm$ 0.07 & \textbf{6.63} & \textbf{348}\\
\hline
\hline
\multirow{2}{*}{\rotatebox[origin=c]{90}{5\%}}
& Baseline VITS & 2.88 $\pm$ 0.11 & 7.40 & 407 \\
\cline{2-5}
& VITS w/ XPB & \textbf{3.22} $\pm$ 0.11 &  \textbf{7.15} & \textbf{383} \\
\hline
\end{tabular}
\label{tab:mainresults_en}
\end{table}

\begin{table}[!t]
\centering
\caption{Obtained results on the Vietnamese test set (here, each MOS score difference between two models is significant with p-value $<$ 0.05). }
\setlength{\tabcolsep}{0.3em}
\begin{tabular}{r|r|c|c|c}
\hline
\multicolumn{2}{c|}{\textbf{Model}} & \textbf{MOS} ($\uparrow$) & \textbf{MCD} ($\downarrow$) & \textbf{RMSE$_{\text{F}_\text{0}}$} ($\downarrow$)\\
\hline
\multicolumn{2}{r|}{Ground truth}    
& 4.26 $\pm$ 0.06 & 0.00 & 0.00\\
\hline
\hline
\multirow{2}{*}{\rotatebox[origin=c]{90}{100\%}}
& Baseline VITS & 3.74 $\pm$ 0.08 & 5.41 & 249 \\
\cline{2-5}
& VITS w/ XPB & \textbf{3.89} $\pm$ 0.08 & \textbf{5.12} & \textbf{234} \\
\hline
\hline
\multirow{2}{*}{\rotatebox[origin=c]{90}{5\%}}
& Baseline VITS &  1.59 $\pm$ 0.05 & 6.20 & 291\\
\cline{2-5}
& VITS w/ XPB   & \textbf{3.35} $\pm$ 0.10 & \textbf{5.39} & \textbf{248}\\
\hline
\end{tabular}
\label{tab:mainresults_vi}
\end{table}

\section{Main results}
\label{sec:result}

Tables \ref{tab:mainresults_en} and \ref{tab:mainresults_vi} show obtained results for English and Vietnamese, respectively. We find that our XPhoneBERT helps improve the performance of VITS on all three evaluation metrics for both English and Vietnamese in both experimental settings. For example, for the first setting of using the whole TTS training set for training, the MOS score significantly increases from 4.00 to 4.14 (+0.14 absolute improvement) for English and from 3.74 to 3.89 (+0.15) for Vietnamese. When it comes to the second setting of using limited training data, XPhoneBERT helps produce larger absolute MOS improvements than those for the first setting. That is, MOS increases from 2.88 to 3.22 (+0.34) for English and especially from 1.59 to 3.35 (+1.76) for Vietnamese, clearly showing the effectiveness of XPhoneBERT. 

Similar to \cite{SaekiTY22}, we also find that the subjective evaluation metric MOS is not ``always'' correlated with the objective evaluation metrics MCD and RMSE$_{\text{F}_\text{0}}$. That is, for Vietnamese in Table \ref{tab:mainresults_vi}, the baseline VITS under the first setting obtains higher MOS but slightly poorer MCD and RMSE$_{\text{F}_\text{0}}$ than the VITS extended with XPhoneBERT under the second setting (MOS: 3.74 vs. 3.35; MCD: 5.41 vs. 5.39; RMSE$_{\text{F}_\text{0}}$: 249 vs. 248).

\begin{figure}[!t]
    \centering
    \includegraphics[width=0.98\linewidth,height=1.4cm]{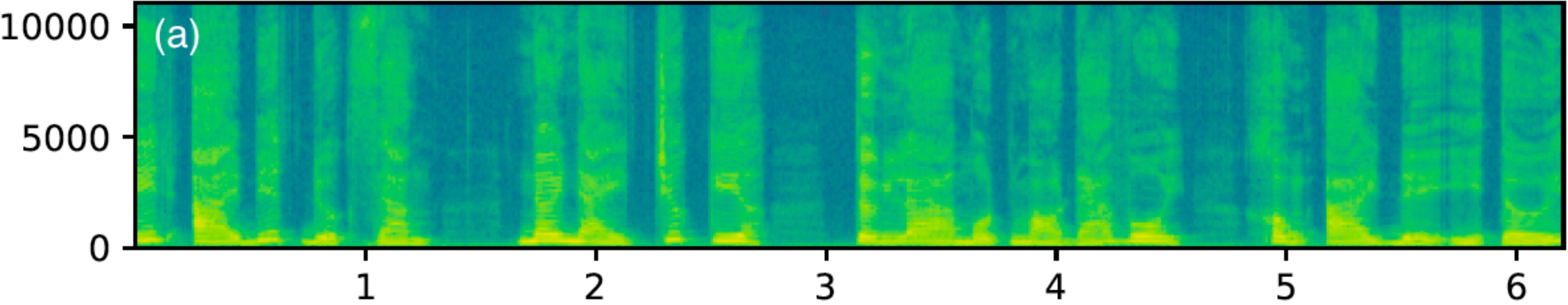}
    \includegraphics[width=0.99\linewidth,height=1.4cm]{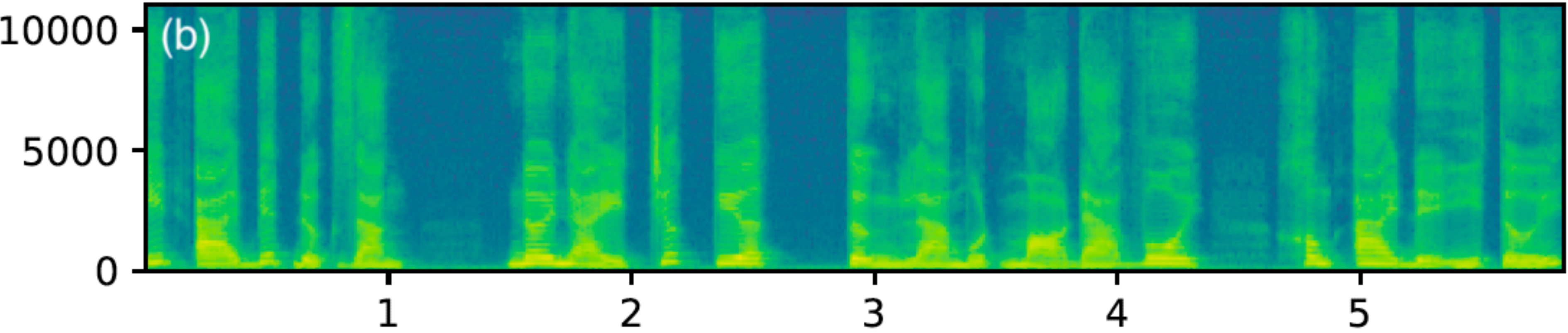}
    \includegraphics[width=0.99\linewidth,height=1.4cm]{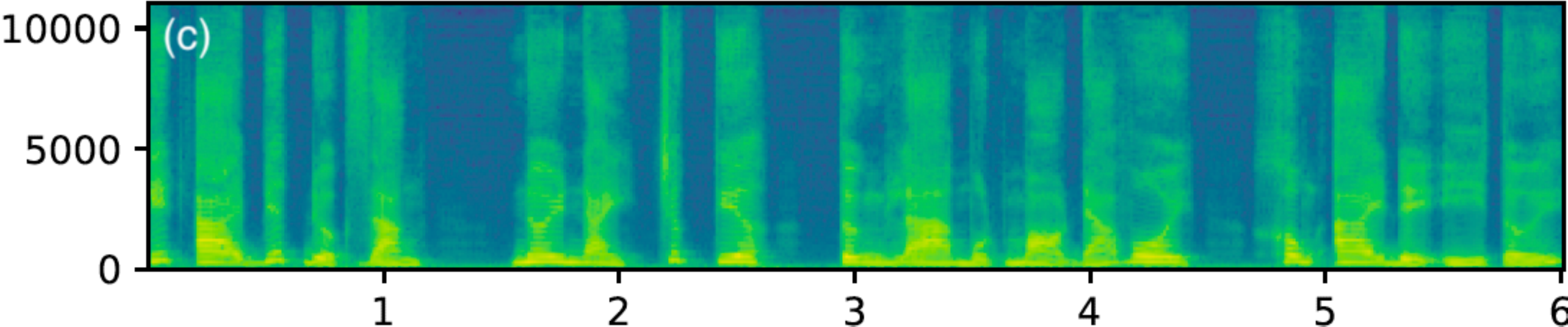}
    \includegraphics[width=0.99\linewidth,height=1.4cm]{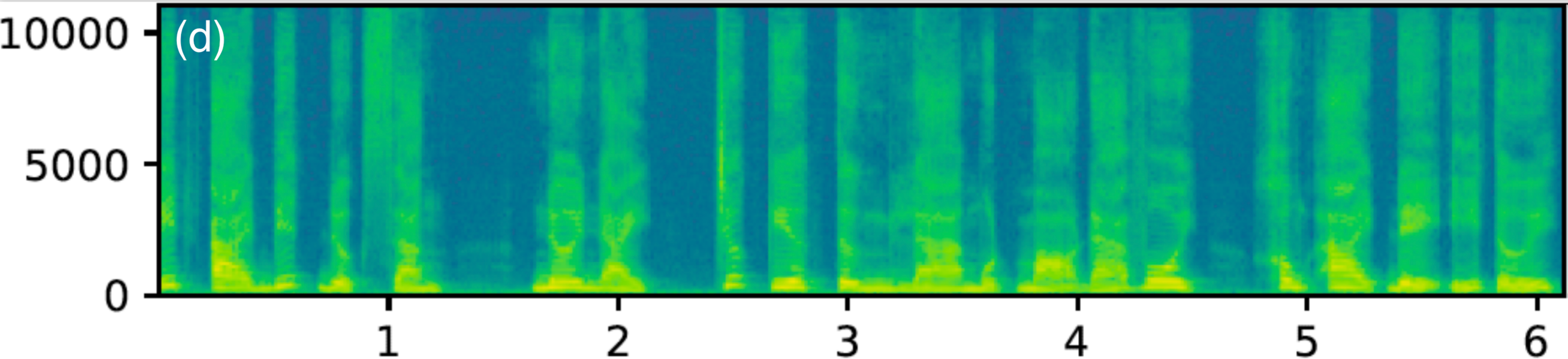}
    \includegraphics[width=0.97\linewidth,height=1.4cm]{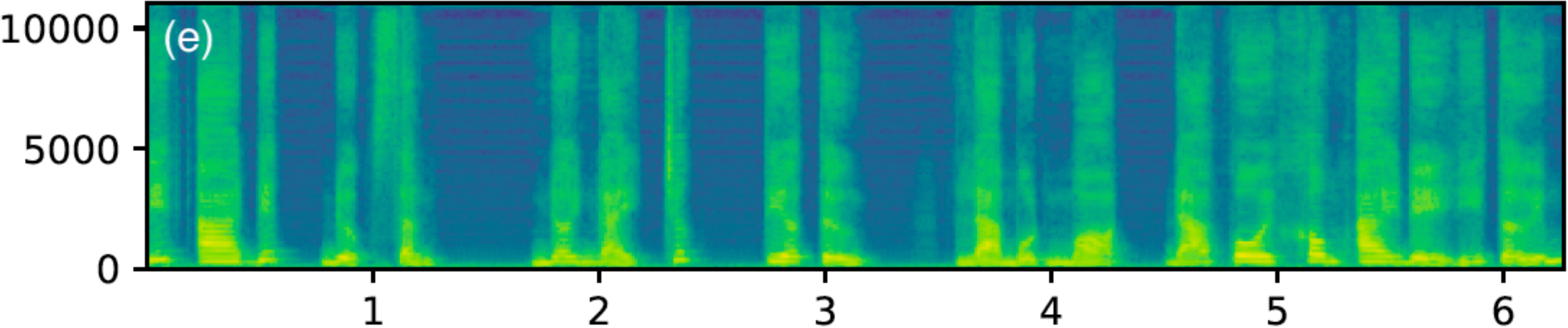}
    \caption{Spectrograms visualization by different models. The text of the speech is ``Ít ai biết được rằng nơi này trước kia từng là một mỏ đá vôi không ai để ý tới'' (Little is known that this place was once a limestone quarry that no one paid any attention to). (a): Ground truth; (b): VITS with XPhoneBERT, under the first experimental setting; (c): VITS with XPhoneBERT, under the second experimental setting; (d): Original VITS, under the first setting; (e): Original VITS, under the second setting.}
    \label{fig:spectrogram}
\end{figure}

From obtained results for the baseline VITS under the first setting and the VITS extended with XPhoneBERT under the second setting in both Tables \ref{tab:mainresults_en} and \ref{tab:mainresults_vi}, we might consider that XPhoneBERT helps synthesize fairly high-quality speech with limited training data. We also visualize the spectrograms of synthesized and ground truth speeches for a Vietnamese text transcription in Figure \ref{fig:spectrogram}, illustrating that XPhoneBERT helps improve the spectral details of the baseline  VITS's output. 

\section{Conclusion}
\label{sec:conclusion}

We have presented the first large-scale multilingual language model XPhoneBERT pre-trained for phoneme representations. We demonstrate the usefulness of XPhoneBERT by showing that using XPhoneBERT as an input phoneme encoder improves the quality of the speech synthesized by a strong neural TTS baseline. XPhoneBERT also helps produce fairly high-quality speech when the training data is limited. We publicly release XPhoneBERT and hope that it can foster future speech synthesis research and applications for nearly 100 languages and locales.

\bibliography{REFs}

\begin{thebibliography}{10}
\providecommand{\url}[1]{#1}
\csname url@samestyle\endcsname
\providecommand{\newblock}{\relax}
\providecommand{\bibinfo}[2]{#2}
\providecommand{\BIBentrySTDinterwordspacing}{\spaceskip=0pt\relax}
\providecommand{\BIBentryALTinterwordstretchfactor}{4}
\providecommand{\BIBentryALTinterwordspacing}{\spaceskip=\fontdimen2\font plus
\BIBentryALTinterwordstretchfactor\fontdimen3\font minus
  \fontdimen4\font\relax}
\providecommand{\BIBforeignlanguage}[2]{{%
\expandafter\ifx\csname l@#1\endcsname\relax
\typeout{** WARNING: IEEEtran.bst: No hyphenation pattern has been}%
\typeout{** loaded for the language `#1'. Using the pattern for}%
\typeout{** the default language instead.}%
\else
\language=\csname l@#1\endcsname
\fi
#2}}
\providecommand{\BIBdecl}{\relax}
\BIBdecl

\bibitem{kong2020hifi}
J.~Kong, J.~Kim, and J.~Bae, ``{HiFi-GAN: Generative Adversarial Networks for
  Efficient and High Fidelity Speech Synthesis},'' in \emph{Proceedings of
  NeurIPS}, 2020.

\bibitem{ren2019fastspeech}
Y.~Ren, Y.~Ruan, X.~Tan, T.~Qin, S.~Zhao, Z.~Zhao, and T.-Y. Liu,
  ``{FastSpeech: Fast, Robust and Controllable Text to Speech},'' in
  \emph{Proceedings of NeurIPS}, 2019.

\bibitem{popov2021grad}
V.~Popov, I.~Vovk, V.~Gogoryan, T.~Sadekova, and M.~Kudinov, ``{Grad-TTS: A
  Diffusion Probabilistic Model for Text-to-Speech},'' in \emph{Proceedings of
  ICML}, 2021, pp. 8599--8608.

\bibitem{tan2022naturalspeech}
X.~Tan, J.~Chen \emph{et~al.}, ``{NaturalSpeech: End-to-End Text to Speech
  Synthesis with Human-Level Quality},'' \emph{arXiv preprint}, vol.
  arXiv:2205.04421, 2022.

\bibitem{shen2018natural}
J.~Shen, R.~Pang \emph{et~al.}, ``{Natural TTS Synthesis by Conditioning
  Wavenet on MEL Spectrogram Predictions},'' in \emph{Proceedings of ICASSP},
  2018, pp. 4779--4783.

\bibitem{li2019neural}
N.~Li, S.~Liu, Y.~Liu, S.~Zhao, and M.~Liu, ``{Neural Speech Synthesis with
  Transformer Network},'' in \emph{Proceedings of AAAI}, 2019, pp. 6706--6713.

\bibitem{kim2020glow}
J.~Kim, S.~Kim, J.~Kong, and S.~Yoon, ``{Glow-TTS: A Generative Flow for
  Text-to-Speech via Monotonic Alignment Search},'' in \emph{Proceedings of
  NeurIPS}, 2020.

\bibitem{renfastspeech}
Y.~Ren, C.~Hu, X.~Tan, T.~Qin, S.~Zhao, Z.~Zhao, and T.-Y. Liu, ``{FastSpeech
  2: Fast and High-Quality End-to-End Text to Speech},'' in \emph{Proceedings
  of ICLR}, 2021.

\bibitem{kim2021conditional}
J.~Kim, J.~Kong, and J.~Son, ``{Conditional Variational Autoencoder with
  Adversarial Learning for End-to-End Text-to-Speech},'' in \emph{Proceedings
  of ICML}, 2021, pp. 5530--5540.

\bibitem{devlin-etal-2019-bert}
J.~Devlin, M.-W. Chang, K.~Lee, and K.~Toutanova, ``{{BERT}: Pre-training of
  Deep Bidirectional Transformers for Language Understanding},'' in
  \emph{Proceedings of NAACL}, 2019, pp. 4171--4186.

\bibitem{RoBERTa}
Y.~Liu, M.~Ott, N.~Goyal, J.~Du, M.~Joshi, D.~Chen, O.~Levy, M.~Lewis,
  L.~Zettlemoyer, and V.~Stoyanov, ``{RoBERTa: {A} Robustly Optimized {BERT}
  Pretraining Approach},'' \emph{arXiv preprint}, vol. arXiv:1907.11692, 2019.

\bibitem{LanCGGSS20}
Z.~Lan, M.~Chen, S.~Goodman, K.~Gimpel, P.~Sharma, and R.~Soricut, ``{ALBERT:}
  {A} lite {BERT} for self-supervised learning of language representations,''
  in \emph{Proceedings of ICLR}, 2020.

\bibitem{hayashi19_interspeech}
T.~Hayashi, S.~Watanabe, T.~Toda, K.~Takeda, S.~Toshniwal, and K.~Livescu,
  ``{Pre-Trained Text Embeddings for Enhanced Text-to-Speech Synthesis},'' in
  \emph{Proceedings of INTERSPEECH}, 2019, pp. 4430--4434.

\bibitem{kenter2020improving}
T.~Kenter, M.~Sharma, and R.~Clark, ``{Improving the Prosody of RNN-based
  English Text-To-Speech Synthesis by Incorporating a BERT model},'' in
  \emph{Proceedings of INTERSPEECH}, 2020, pp. 4412--4416.

\bibitem{xu2021improving}
G.~Xu, W.~Song, Z.~Zhang, C.~Zhang, X.~He, and B.~Zhou, ``{Improving Prosody
  Modelling with Cross-Utterance BERT Embeddings for End-to-end Speech
  Synthesis},'' in \emph{Proceedings of ICASSP}, 2021, pp. 6079--6083.

\bibitem{jia2021png}
Y.~Jia, H.~Zen, J.~Shen, Y.~Zhang, and Y.~Wu, ``{PnG BERT: Augmented BERT on
  Phonemes and Graphemes for Neural TTS},'' in \emph{Proceedings of
  INTERSPEECH}, 2021, pp. 151--155.

\bibitem{zhang2022mixed}
G.~Zhang, K.~Song, X.~Tan, D.~Tan, Y.~Yan, Y.~Liu, G.~Wang, W.~Zhou, T.~Qin,
  T.~Lee, and S.~Zhao, ``{Mixed-Phoneme BERT: Improving BERT with Mixed Phoneme
  and Sup-Phoneme Representations for Text to Speech},'' in \emph{Proceedings
  of INTERSPEECH}, 2022, pp. 456--460.

\bibitem{phoneme-levelBERT}
Y.~A. Li, C.~Han, X.~Jiang, and N.~Mesgarani, ``{Phoneme-Level BERT for
  Enhanced Prosody of Text-to-Speech with Grapheme Predictions},'' \emph{arXiv
  preprint}, vol. arXiv:2301.08810, 2023.

\bibitem{donlpotherlanguages}
\BIBentryALTinterwordspacing
S.~Ruder, ``{Why You Should Do NLP Beyond English},''
  https://ruder.io/nlp-beyond-english/, 2020. [Online]. Available:
  \url{https://ruder.io/nlp-beyond-english/}
\BIBentrySTDinterwordspacing

\bibitem{NIPS2017_7181}
A.~Vaswani, N.~Shazeer, N.~Parmar, J.~Uszkoreit, L.~Jones, A.~N. Gomez,
  {\L}.~Kaiser, and I.~Polosukhin, ``{Attention is All you Need},'' in
  \emph{Proceedings of NIPS}, 2017, pp. 5998--6008.

\bibitem{zhu2022charsiu-g2p}
J.~Zhu, C.~Zhang, and D.~Jurgens, ``{ByT5 model for massively multilingual
  grapheme-to-phoneme conversion},'' in \emph{Proceedings of INTERSPEECH},
  2022, pp. 446--450.

\bibitem{guo-etal-2020-wiki}
M.~Guo, Z.~Dai, D.~Vrande{\v{c}}i{\'c}, and R.~Al-Rfou, ``{{W}iki-40{B}:
  Multilingual Language Model Dataset},'' in \emph{Proceedings of LREC}, 2020,
  pp. 2440--2452.

\bibitem{wikidump}
\BIBentryALTinterwordspacing
W.~Foundation. Wikimedia downloads. [Online]. Available:
  \url{https://dumps.wikimedia.org}
\BIBentrySTDinterwordspacing

\bibitem{lhoest-etal-2021-datasets}
Q.~Lhoest, A.~Villanova~del Moral, Y.~Jernite \emph{et~al.}, ``{Datasets: A
  Community Library for Natural Language Processing},'' in \emph{Proceedings of
  EMNLP: System Demonstrations}, 2021, pp. 175--184.

\bibitem{nguyen-etal-2018-fast}
D.~Q. Nguyen, D.~Q. Nguyen, T.~Vu, M.~Dras, and M.~Johnson, ``{A Fast and
  Accurate Vietnamese Word Segmenter},'' in \emph{Proceedings of LREC}, 2018,
  pp. 2582--2587.

\bibitem{vu-etal-2018-vncorenlp}
T.~Vu, D.~Q. Nguyen, D.~Q. Nguyen, M.~Dras, and M.~Johnson, ``{VnCoreNLP: A
  Vietnamese Natural Language Processing Toolkit},'' in \emph{Proceedings of
  NAACL: Demonstrations}, 2018, pp. 56--60.

\bibitem{nemo}
O.~Kuchaiev, J.~Li \emph{et~al.}, ``{NeMo: a toolkit for building AI
  applications using Neural Modules},'' in \emph{Proceedings of NeurIPS
  Workshop on Systems for ML}, 2019.

\bibitem{Bernard2021}
M.~Bernard and H.~Titeux, ``{Phonemizer: Text to Phones Transcription for
  Multiple Languages in Python},'' \emph{Journal of Open Source Software},
  vol.~6, no.~68, p. 3958, 2021.

\bibitem{ott2019fairseq}
M.~Ott, S.~Edunov, A.~Baevski, A.~Fan, S.~Gross, N.~Ng, D.~Grangier, and
  M.~Auli, ``{fairseq: A Fast, Extensible Toolkit for Sequence Modeling},'' in
  \emph{Proceedings of NAACL-HLT 2019: Demonstrations}, 2019, pp. 48--53.

\bibitem{KingmaB14}
D.~P. Kingma and J.~Ba, ``{Adam: A Method for Stochastic Optimization},'' in
  \emph{Proceedings of ICLR}, 2015.

\bibitem{wolf-etal-2020-transformers}
T.~Wolf, L.~Debut \emph{et~al.}, ``{Transformers: State-of-the-Art Natural
  Language Processing},'' in \emph{Proceedings of EMNLP 2020: System
  Demonstrations}, 2020, pp. 38--45.

\bibitem{ljspeech17}
K.~Ito and L.~Johnson, ``{The LJ Speech Dataset},''
  \url{https://keithito.com/LJ-Speech-Dataset/}, 2017.

\bibitem{phobert}
D.~Q. Nguyen and A.~T. Nguyen, ``{PhoBERT: Pre-trained language models for
  Vietnamese},'' in \emph{Findings of EMNLP}, 2020, pp. 1037--1042.

\bibitem{loshchilov2018decoupled}
I.~Loshchilov and F.~Hutter, ``{Decoupled Weight Decay Regularization},'' in
  \emph{Proceedings of ICLR}, 2019.

\bibitem{tamamori2017speaker}
A.~Tamamori, T.~Hayashi, K.~Kobayashi, K.~Takeda, and T.~Toda,
  ``{Speaker-dependent WaveNet vocoder},'' in \emph{Proceedings of
  INTERSPEECH}, 2017, pp. 1118--1122.

\bibitem{SaekiTY22}
T.~Saeki, K.~Tachibana, and R.~Yamamoto, ``{DRSpeech: Degradation-Robust
  Text-to-Speech Synthesis with Frame-Level and Utterance-Level Acoustic
  Representation Learning},'' in \emph{Proceedings of INTERSPEECH}, 2022, pp.
  793--797.

\end{thebibliography}
\bibliographystyle{IEEEtran}

\end{document}